\documentclass[letterpaper]{article} 
\usepackage{aaai23}  
\usepackage{times}  
\usepackage{helvet}  
\usepackage{courier}  
\usepackage[hyphens]{url}  
\usepackage{graphicx} 
\usepackage{natbib}  
\usepackage{caption} 
\usepackage{algorithm}
\usepackage{algorithmic}

\usepackage{newfloat}
\usepackage{listings}
\DeclareCaptionStyle{ruled}{labelfont=normalfont,labelsep=colon,strut=off} 
\floatstyle{ruled}
\newfloat{listing}{tb}{lst}{}
\floatname{listing}{Listing}
\usepackage{url} 
\usepackage{bm}
\usepackage{multirow}
\usepackage{booktabs}
\usepackage{cite}
\usepackage{subfigure}
\usepackage{tikz}
\usepackage{comment} 
\usepackage{amsmath,amssymb}
\usepackage{color}
\usepackage{tabu}
\usepackage{multirow}
\usepackage{booktabs}
\usepackage{xspace}
\usepackage{cite}
\usepackage[algo2e,ruled,linesnumbered]{algorithm2e}
\usepackage{amsthm}
\newtheorem{theorem}{Theorem}

\usepackage{xcolor}
\definecolor{codegreen}{rgb}{0,0.5,0}
\definecolor{codeblue}{rgb}{0.5,0.5,1}
\definecolor{codegray}{rgb}{0.6,0.6,0.6}
\usepackage{framed}

\title{FAKD: Feature Augmented Knowledge Distillation for Semantic Segmentation }
\author{
    Jianlong Yuan$^1$,
    Qian Qi$^1$,
    Fei Du$^1$,
    Zhibin Wang$^1$,
    Fan Wang$^1$,
    Yifan Liu$^2$
}
\affiliations{
    \textsuperscript{\rm 1}Alibaba Group\\
    \textsuperscript{\rm 2}University of Adelaide\\
    \{gongyuan.yjl, qi.qian, dufei.df, zhibin.waz, fan.w\}@alibaba-inc.com\\
    yifan.liu04@adelaide.edu.au
}

\usepackage{bibentry}

\begin{document}

\maketitle

\begin{abstract}
In this work, we explore data augmentations for knowledge distillation on semantic segmentation. To avoid over-fitting to the noise in the teacher network, a large number of training examples is essential for knowledge distillation. Image-level argumentation techniques like flipping, translation or rotation are widely used in previous knowledge distillation framework. Inspired by the recent progress on semantic directions on feature-space, we propose to include augmentations in feature space for efficient distillation. Specifically, given a semantic direction, an infinite number of augmentations can be obtained for the student in the feature space. Furthermore, the analysis shows that those augmentations can be optimized simultaneously by minimizing an upper bound for the losses defined by augmentations. Based on the observation, a new algorithm is developed for knowledge distillation in semantic segmentation. Extensive experiments on four semantic segmentation benchmarks demonstrate that the proposed method can boost the performance of current knowledge distillation methods without any significant overhead. Code is available at: \url{https://github.com/jianlong-yuan/FAKD}.
\end{abstract}

\section{Introduction}
\label{sec:intro}
Semantic segmentation/Pixel labeling is a fundamental problem in computer vision, which is widely used in scene parsing, human body parsing, and many downstream applications. 
It is a per-pixel classification problem that assigns each pixel in an image to a specific set of predefined classes.
With the development of deep learning, semantic segmentation has made tremendous progress in recent years and achieved impressive results on large benchmark datasets \cite{Everingham10, mottaghi_cvpr14, cordts2016cityscapes, zhou2017scene}. 
However, advanced segmentation models usually require complicated model designs and expensive computations. As a result, it limits the potential for computationally constrained applications on devices.

To alleviate the computational cost, some efforts have been devoted to designing lightweight networks, especially for semantic segmentation \cite{yu2018bisenet, mehta2018espnet, zhao2018icnet, yu2021bisenet, chen2019fasterseg, li2019dfanet}. In addition, knowledge distillation (KD) attracts attention to train lightweight semantic segmentation networks effectively by leveraging the knowledge from the teacher network. 
The conventional KD method for the classification problem is to apply a Kullback–Leibler (KL) divergence between the output of the teacher and the student network for each example. 
However, mimicking every pixel's class distribution is inappropriate for semantic segmentation since the noise accumulated from the pixel-level activation may degenerate the performance. 
SKDS \cite{liu2019structured, liu2020structured}  aggregates a sub-set of different spatial locations according pair-wise relations. Moreover, IFVD \cite{hou2020inter} exploits the inter-class relations among different locations. CWD~\cite{shu2021channel} proposes channel-wise knowledge distillation by minimizing the channel distribution, indicating the spatial location for each class. CIRKD~\cite{yang2022cross} focuses on transferring structured pixel-to-pixel and pixel-to-region relations among the whole images.

However, the teacher and student can have different training procedures, which makes the transfer challenging even with the sophisticated objective function for semantic segmentation. Moreover, optimization based on a limited number of training examples can make the knowledge transfer for the whole data distribution more challenging. To capture the data distribution on unseen data, augmentation is a prevalent strategy to improve the generalization of deep models. Concretely, image-level data augmentation operators, including random crop, random scale, and flip~\cite{he2016deep}, are widely applied in existing methods. Apparently, the number of applicable augmentation operators for each image is limited, which may be insufficient for knowledge distillation that has to mimic the diverse patterns from the teacher. Moreover, this kind of image space based augmentations has to be fed forward from the input to the whole network, thus only a limited number of augmentations can be adopted for optimization.

Recently, some works~\cite{upchurch2017deep,li2016convolutional, bengio2013better, wang2019implicit, wang2021regularizing} propose there are many semantic directions in the deep feature space, such that translating a data sample along one of these directions in the feature space produces a feature representation corresponding to another sample with the same class but different semantics.
Meanwhile, compared to the augmentation in raw images, that in high-level features is more flexible and can be obtained along with arbitrary semantic directions.
Therefore, 
motivated by these,
we propose to augment the samples in the feature space to apply sufficient and diverse augmentations for knowledge distillation.

Specifically, the direction of intra-class variation that contains the semantic information of each class is essential and adopted for the perturbation in feature space. Based on the intra-class variance, we augment examples in feature space according to the corresponding Gaussian distribution. Thanks to this procedure, we can have an infinite number of augmentations from feature space, and an upper bound of the loss function can be optimized accordingly to transfer the knowledge from the teacher sufficiently. 
With massive augmented examples for the student, the nice properties, e.g., large margin, from the teacher can be preserved for generalization. 

Finally, based on the distribution definition, different knowledge distillation objective functions will have different surrogate functions for the upper bounds of infinite augmentations. 
In this work, two types of Kullback–Leibler divergence-based knowledge distillation loss are investigated, and the corresponding loss functions are derived. Experiment results show that the proposed feature augmentation method can improve the performance over baseline KL–based knowledge distillation losses on four different datasets.

Concretely, our contributions are three-fold:
\begin{itemize}
    \item We propose a novel feature augmented knowledge distillation (FAKD) method for semantic segmentation. Specifically, the student model will mimic the teacher model with infinite augmented samples. To the best of our knowledge, it is the first time that the feature-level augmentation method has been explored in knowledge distillation for semantic segmentation.
    \item We provide a theoretical analysis from the infinite number of augmentations in feature space to the upper-bound loss, so as to theoretically demonstrate our proposal for different knowledge distillation methods.
    \item We demonstrate the effectiveness of our approach applying various network structures on four benchmark datasets: ADE20K, Pascal Context, Cityscapes, and Pascal VOC.
\end{itemize}

\section{Related Work}
\label{sec:related work}
\subsection{Semantic Segmentation}
Semantic segmentation can be regarded as a pixel-level classification problem. The fully convolutional network(FCN)~\cite{long2015fully} is a pioneering work in the field of semantic segmentation, which can adapt to any scale input. To improve the performance of the segmentation network, many researchers try to improve FCN in different ways. 
In \cite{chen2017deeplab, yu2015multi, zhao2017pyramid, yang2018denseaspp, peng2017large, chen2018encoder, yuan2020multi}, the receptive field is enlarged to capture more details. In \cite{zhang2018context, he2019adaptive, lin2017refinenet, zhou2019context, yuan2020object, yuan2018ocnet, yu2020context}, the contextual information is combined to improve the semantic understanding of the model. In~\cite{ding2019boundary, bertasius2016semantic, yuan2020segfix, zhen2020joint, takikawa2019gated, yu2018learning, chen2016semantic}, the boundary information is considered to boost the segmentation accuracy further. In \cite{wang2018non, fu2019dual, zhong2020squeeze, huang2019ccnet, zhao2018psanet}, some new attention modules are designed to enrich the representations of the feature map. 
Moreover, \cite{xie2021segformer, zheng2021rethinking, wang2021end} introduce transformer blocks into the Semantic Segmentation task, which boosts the performance with a large margin. 
Although the state-of-the-art performance keeps improving nowadays, the models have become larger than before. The high computational cost limits the application of the segmentation models over resource-limited mobile devices.

Efficient segmentation networks attract wide attention due to the need for real-time inference. Most of the works pay attention to designing lightweight networks with cheap operations. ENet \cite{paszke2016enet} is a efficient segmentation network with  early downsampling, lightweight decoder. ESPNet \cite{mehta2018espnet} introduces more cheap spatial pyramid of dilated convolution. ICNet \cite{zhao2018icnet} proposes a cascade structure to use low-resolution and high-resolution features. BiSeNet \cite{yu2018bisenet} combines a spatial path and a context path to process features efficiently.

\subsection{Knowledge distillation} 
Knowledge distillation has been extensively studied in recent years. The core idea of knowledge distillation is to transfer meaningful knowledge from a cumbersome teacher into a smaller and faster student. Most knowledge distillation methods for image classification networks could be classified into three classes, probability-based, feature-based and relation-based methods. \cite{hinton2015distilling, zhou2021rethinking} distill the output logits of the teacher network as soft labels to the student. Feature-based knowledge distillation methods \cite{romero2014fitnets, heo2019comprehensive, zagoruyko2016paying} focuses on the feature maps.  Finally, relation-based KD~\cite{peng2019correlation, park2019relational,qian2022improved} aligns correlations among multiple instances between the student and teacher networks. However, these image-level KD methods are not suitable for pixel-level semantic segmentation.

Some works pay attention to utilizing knowledge distillation to semantic segmentation.  The pioneering work \cite{liu2019structured, liu2020structured} proposes Structured Knowledge Distillation (SKDS), which enables the output of the student model to transfer the structure knowledge among pixels from the teacher model to the student model through pairwise relationships and adversarial training. \cite{wang2020intra} focuses on intra-class feature variations between pixels with the same label, where the set of cosine distances between the features of each pixel and its corresponding class prototype is constructed to convey structural knowledge. \cite{shu2021channel} proposes channel-wise distillation to guide the student to mimic the teacher's channel-wise distribution, which indicates the spatial location for each class. 
\cite{yang2022cross} focuses on transferring structured pixel-to-pixel and pixel-to-region relations among the whole images.
These methods try to develop objective functions to align useful knowledge between the teacher and the student network. 
Our approach, however, is to face how effective knowledge distillation can be performed when teacher and student training procedures are not aligned in real-life scenarios.

\begin{figure*}[!htbp]
  \centering
  \includegraphics[width=\linewidth]{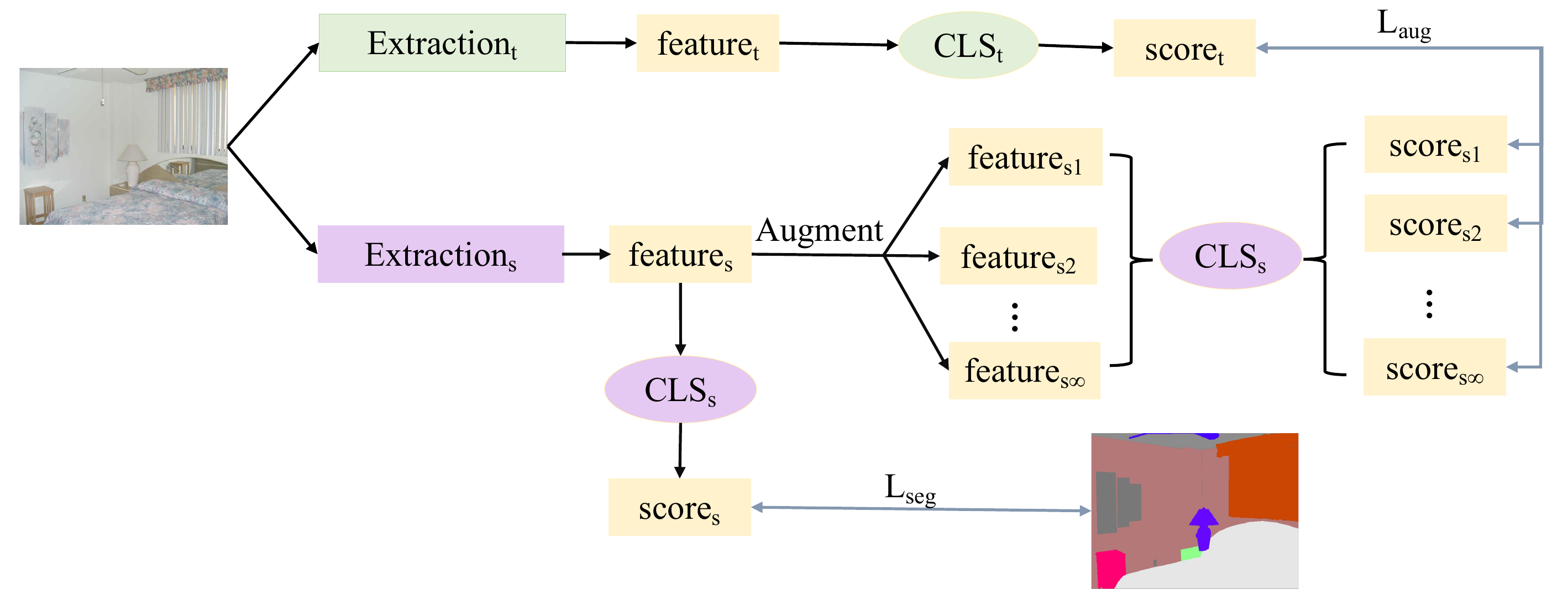}
  \caption{Illustration of the proposed feature augmented knowledge distillation (FAKD). The parameters of the teacher network are fixed during training. The feature of the student network is augmented to mimic the effect of infinite samples. The weight for the `CLS$_s$' is shared.}
  \label{fig:framework}
  \vspace{-1em}
\end{figure*}

\section{Methods}
\label{sec:methods}

The overall pipeline of the proposed method, feature augmented knowledge distillation (FAKD), is illustrated in Figure \ref{fig:framework}. We will first revisit two Kullback–Leibler (KL) divergence-based knowledge distillation losses and then investigate the new loss functions with an infinite number of augmentations.

\subsection{Revisit Knowledge Distillation}
\label{sec: Sketch of Knowledge Distillation}
KL divergence is a standard metric to evaluate the difference between distributions, which is also widely used in designing knowledge distillation objective functions. In this subsection, we will revisit two popular knowledge distillation objective functions, i.e., pixel-wise distillation~\cite{hinton2015distilling} and channel-wise distillation~\cite{shu2021channel}. 

The pixel-wise distillation (PD) is straightforward that mimics the output from the teacher for each pixel. It directly extends the knowledge distillation protocol for classification to semantic segmentation and is widely used as a baseline method in existing semantic segmentation distillation works~\cite{liu2019structured, hou2020inter}. The channel-wise distillation (CWD)~\cite{shu2021channel}  aligns the output distribution for each channel rather than the pixel. It achieves state-of-the-art performance on benchmark semantic segmentation datasets.

Let $s_{i} \in \mathbb{R}^{A}$ and $t_{i} \in \mathbb{R}^{A}$ be the feature of the student and teacher of the $i$ pixel. $A$ is the number of the feature channel before the final classifier. A linear classifier is usually applied to the feature to get the final logit, for example for the teacher's output of the pixel $i$ for the $c$ class, $q^{c}_{t_{i}} = \omega_{c}^{\top}t_{i}+b_{c}$. 
The loss function for the pixel-wise distillation~\cite{hinton2015distilling, liu2019structured, liu2020structured} can be written as
\begin{multline}
\ell_{PD} = \frac{1}{M} \sum_{i=1}^{M} \sum_{c=1}^{C} \\
\left(-\frac{e^{\omega_{c}^{\top}t_{i}+b_{c}}}{{\sum_{k=1}^{C}e^{{\omega_{k}^{\top}t_{k}+b_{k}}}}}  
\log \left ( \frac{e^{\omega_{c}^{\top}s_{i}+b_{c}}}{{\sum_{k=1}^{C}e^{{\omega_{k}^{\top}s_{k}+b_{k}}}}} \right) \right),
\label{eq:pd_detail}
\end{multline}
where $M$ donates the number of pixels and $C$ is the number of categories. $W =[\omega_{1}, \dots , \omega_{C}]^\top \in \mathbb{R}^{C \times A} $ and $B =[b_{1}, \dots , b_{C}]^\top \in \mathbb{R}^{C} $ are the classification convolution weights and bias, respectively. The softmax operator is applied to every pixel prediction for all classes to find out the class distribution for all pixels. Then the KL divergence between two class distributions will be minimized by $\ell_{PD}$.

For CWD~\cite{shu2021channel}, it aims to minimize the KL divergence between the channel-wise probability map of the two networks, which can be written as
\begin{multline}
\ell_{CWD} = \frac{\tau^2}{C} \sum_{i=1}^{M} \sum_{c=1}^{C} \\ \left(-\frac{e^{\frac{\omega_{c}^{\top}t_{i}+b_{c}}{\tau}}}{{\sum_{k=1}^{M}e^{{\frac{\omega_{c}^{\top}t_{k}+b_{c}}{\tau}}}}} 
\log \left(\frac{e^{\frac{\omega_{c}^{\top}s_{i}+b_{c}}{\tau}}}{{\sum_{k=1}^{M}e^{{\frac{\omega_{c}^{\top}s_{k}+b_{c}}{\tau}}}}}\right)\right)
\label{eq:cwd}
\end{multline}
where $\tau$ is the temperature factor. Note that the main difference between Eq.~\ref{eq:cwd} and Eq.~\ref{eq:pd_detail} lies in the softmax operator. Different normalization operators leverage different distributions.

\subsection{Feature Augmented Knowledge Distillation}
\label{sec: Semantic-Wise Knowledge Distillation}

To improve the training effectiveness, the feature from the student $s_i$ can be perturbed $N$ times while all of these augmented examples should be aligned with the feature from the teacher, i.e., $t_i$. Consequently, the distillation loss can be optimized with the set of pairs of $\{(s_{i}^{1}, t_{i}),\dots ,(s_{i}^{N}, t_{i}) \}_{i=1}^{M}$. We will have the analysis for CWD while a similar analysis can be conducted on PD.

Let $p_{t_{i}}^{c}$ donate the output from the teacher.
\[p_{t_{i}}^{c} = \frac{e^{\frac{\omega_{c}^{\top}t_{i}+b_{c}}{\tau}}}{{\sum_{k=1}^{M}e^{{\frac{\omega_{c}^{\top}t_{k}+b_{c}}{\tau}}}}}\]
The CWD loss in Eq.~\ref{eq:cwd} can be rewritten with the augmented examples.
\begin{equation}
    \frac{ \tau^2}{C} \sum_{i=1}^M \sum_{c=1}^C  \frac{1}{N} \sum_{n=1}^N - p_{t_{i}}^{c} \log \left( {\frac{e^{\frac{ \omega_{c}^\top s_{i}^n+b_{c}}{\tau }}}{\sum_{k=1}^M {e^{\frac{\omega_{c}^\top s_{k}^n+b_c}{\tau }}}}} \right)
\label{eq:swd import}
\end{equation}

With a large $N$, the information from the teacher can be leveraged more sufficiently. Now we consider an extreme case such that $N\to \infty$. With the infinity augmentations, the expectation over the set of augmentations is optimized instead, and the loss function becomes
\begin{equation}
    \ell_{aug}=\frac{ \tau^2}{C} \mathbb{E}_{\{\hat{s}_{j}\}_{j=1}^M}\sum_{i=1}^M \sum_{c=1}^C p_{t_{i}}^{c}  \left( \log \left(\sum_{k=1}^{M}e^{\frac{w_{c}^{\top} (\hat{s}_{i}-\hat{s}_{k})}{\tau} } \right) \right).
\label{eq:swd e}
\end{equation}

The loss can be upper-bounded by an empirical loss as follows.
\begin{theorem}
By assuming that $\hat{s}_i \sim \mathcal{N}(s_{i}, \lambda_{i} \Sigma_{i})$, we have
\begin{align*}
&\ell_{aug}\leq \\
&\frac{ \tau^2}{C} \sum_{i=1}^M \sum_{c=1}^C p_{t_{i}}^{c} \log \left( \sum_{k=1}^{C} \left[ e^{\frac{w_{c}^{\top} (s_{i} - s_{k})}{\tau} + \frac{w_c^\top (\lambda_{c} \Sigma_{i} + \lambda_{k} \Sigma_{k}) w_c}{2\tau}} \right] \right).
\end{align*}
\end{theorem}

According to the Jason's Inequality, the loss in Eq.~\ref{eq:swd e} can be upper-bounded as
\begin{equation}
    \ell_{aug} \le \frac{ \tau^2}{C} \sum_{i=1}^M \sum_{c=1}^C p_{t_{i}}^{c} \log \left(\mathbb{E}\left(\sum_{k=1}^{M}e^{\frac{w_{c}^{\top} (\hat{s}_{i}-\hat{s}_{k})}{\tau} }\right) \right).
\label{eq:swd jason}
\end{equation}

We follow existing work~\cite{gal2015bayesian, kendall2017uncertainties}  and apply Gaussian distribution to approximate the distribution for deep features. Specifically, we assume that $\hat{s}_i \sim \mathcal{N}(s_{i}, \lambda_{i} \Sigma_{i})$, where $\Sigma_{i}$ are the statistics covariance of the semantic distribution of the $i$-th example.
Particularly, following \cite{wang2019implicit}, the covariance is updated from the batch data during the training process
, and more details are in Supplementary. 
With the assumption, $\hat{s}_{i} - \hat{s}_{k}$ also follows the Gaussian Distribution:

\begin{multline}
 \frac{w_{c}^{\top } (\hat{s}_{i} - \hat{s}_{k})}{\tau}
\sim \\
\mathcal{N} \left(\frac{w_{c}^{\top }( s_{i} - s_{k})}{\tau}, \frac{w_c^\top (\lambda_{i} \Sigma_{i} + \lambda_{k} \Sigma_{k}) w_c}{\tau} \right).
\label{eq: gaus}
\end{multline}

For a variable $x$ that follows Gaussian distribution $\mathcal{N}(\mu, \Sigma)$, the moment-generating function shows that $\mathbb{E}[e^{a^{\top} x}] = e^{a^{\top} \mu + \frac{1}{2} a^{\top} \Sigma a } $.
Therefore, by taking the statistics for each example, the upper-bound in Eq.~\ref{eq:swd jason} can be simplified as
\begin{multline}
 \ell_{aug}^{CWD} = \frac{ \tau^2}{C} \sum_{i=1}^M \sum_{c=1}^C p_{t_{i}}^{c} \\
 \log \left( \sum_{k=1}^{C} \left[ e^{\frac{w_{c}^{\top} (s_{i} - s_{k})}{\tau} + 
 \frac{w_c^\top (\lambda_{c} \Sigma_{i} + \lambda_{k} \Sigma_{k}) w_c}{2\tau}} \right] \right) 
\label{equ: swd generating}
\end{multline}

\paragraph{Remark} By minimizing the empirical loss in Eq.~\ref{equ: swd generating}, an infinite number of augmentations will be leveraged for learning the student, which can help capture the knowledge from the teacher. With the same process, the upper-bound for PD with infinite augmentations can be obtained as follows.

\begin{multline}
\ell_{aug}^{PD} = \frac{1}{M} \sum_{i=1}^M \sum_{c=1}^C  
\frac{e^{\omega_{c}^{\top}t_{i}+b_{c}}}{{\sum_{k=1}^{C}e^{{\omega_{k}^{\top}t_{k}+b_{k}}}}}\\
\log \left(\sum_{k=1}^{C} \left [ e^{(w_{k}^{\top } - w_{c}^{\top }) s_{i} + b_{k} -b_{c} + \frac{\lambda }{2} (w_{k}^{\top } - w_{c}^{\top })^{\top} \Sigma_{c} (w_{k}^{\top } - w_{c}^{\top })} \right ] \right) 
\label{equ: swd PD}
\end{multline}

The proposed method is summarized in the Algorithm~\ref{framework}. 
Firstly, a powerful teacher network, a lightweight student network, and covariance matrices are defined.
Teacher and student networks are composed of feature extraction and classification.
Taking a forward as an example, for each mini-batch, the teacher's classification logits, the student's classification logits, and the student's feature before logits are obtained. Then, the extracted features and the classification parameters are used to update the covariance. At the same time, $\lambda$ will be updated with the current number of iterations. Finally, the parameters of the student model will be optimized with the corresponding loss.

\begin{algorithm}
  \caption{The FAKD framework}\label{framework}
    \textbf{Teacher:} Teacher network with extraction function $f_t$ and classification function $g_t$; \\
    \textbf{Student:} Student network with extraction function $f_s$ and classification function $g_s$; \\
    \textbf{Input:} Training image dataset $A$, super parameter $\lambda$ and Covariance $\Sigma$ initialized with zero; \\
    \For{$step=1 ,..., n_{steps}$ }{
        $x, y$=Sample($A$),\\
        $p_t$ = $g_t(f_t(x))$, \\
        $s$ = $f_s(x)$, \\
        $p_s$ = $g_s(s)$, \\
        update $\Sigma$ and $\lambda$ with current $s$, $y$ and $step$, \\
        loss $=$ $\ell_{seg}$($p_{s}$, $y$) $+$ $\ell_{aug}$(t($W_{g_s}$, $\Sigma$, $s$, $p_s$, $\lambda$),$p_t$); \\
        minimizing the loss and update student parameters $\theta_{s}$; \\
    }
\end{algorithm}

\subsection{Experimental Setup}

\noindent\textbf{Dataset.} We employ four popular semantic segmentation datasets to conduct our experiments. ADE20K \cite{zhou2017scene} contains 20k/2k/3k images for train/val/test with 150 semantic classes. Cityscapes \cite{cordts2016cityscapes} is an urban scene parsing dataset that contains 2975/500/1525 finely annotated images used for train/val/test. And the performance is evaluated on 19 classes. Pascal Context \cite{mottaghi_cvpr14} provides dense annotations which contains 4998/5105/9637 train/val/test images. We use 59 object categories for training and testing. Pascal VOC contains 21 classes including 20 object categories and one background class. Following the procedure of \cite{zhao2017pyramid, chen2018encoder} , we use augmented data with annotation of resulting 10582, 1449, and 1456 images for train/val.
Our results are all reported on the validation set. 

\noindent\textbf{Evaluation metrics.} We report the mean Intersection over Union (mIoU) and pixel accuracy (mAcc), a standard metric for semantic segmentation.

\noindent\textbf{Network architectures.} On each dataset, the same series of teacher models and student models are used. 

\noindent\textbf{Implementation details.} Following the standard data augmentation, we employ random flipping, cropping and scaling in the range of [0.5, 2]. 
All experiments are optimized by SGD with a momentum of 0.9, a batch size of 16, and 512 x 512 crop size. We use an initial learning rate of 0.01 for ADE20K, Cityscapes, and Pascal VOC. In addition, we use an initial learning rate of 0.004 for Pascal Context. The number of the total training iterations is 40K. Following previous methods \cite{chen2018encoder, zhao2017pyramid}, we use the poly learning rate policy where current learning rate equals to the base one multiplying $(1 - \frac{iter}{max_{iter}})^{0.9}$. We report the single-scale testing result. To be fair, each method uses the same parameters again for the same data set.

\noindent\textbf{Compared distillation methods}. On each dataset, we compare with state-of-the-art segmentation distillation methods including SKDS~\cite{liu2019structured}, IFVD \cite{hou2020inter} and CWD \cite{shu2021channel}, CIRKD~\cite{yang2022cross}. We re-run SKDS, IFVD, CWD and FAKD on 4 NIVIDIA V100 GPUs. In addition, due to the limitation of GPU memory, we re-run CIRKD on 4 NVIDIA A100. We take same super parameters as CWD in our method. And all of teacher models are from \cite{mmseg2020}.

\subsection{Compared with State-of-the-art Methods}
\noindent\textbf{ADE20K.}
We conduct the comparison experiments with other state-of-the-art algorithms on Table \ref{tab: SOTA on AED20K}. 
We also apply our proposed framework to different teachers following \cite{liu2019structured, wang2020intra, shu2021channel, yang2022cross} to make a fair comparison.
All of the students' backbones except that with $^\dagger$ are initialized with the pre-trained weights on ImageNet classification.
With the help of different teachers, we find that all of the methods improve different student networks. Specifically, our method achieves the best segmentation performance. 
From Table \ref{tab: SOTA on AED20K}, we can see that our method can effectively improve the performance by $1.48\%$, $1.82\%$, $1.1\%$, $1.4\%$, $1.05\%$, respectively. 
Compared with the student, our method can improve the performance by $5.88\%$, $9.85\%$, $3.77\%$, $5.91\%$, $8.06\%$, respectively.
It demonstrates that our method does not depend on a specific model structure. 
Meanwhile, our method can dramatically improve the results when the student model has not learned the knowledge, as PSPnet-R18$^\dagger$. 
From Figure in Supplementary, 
we further show the qualitative segmentation results. Our results have more detailed segmentation results.

\begin{table}[htbp]
\begin{center}
\caption{Performance comparison with state-of-the-art distillation methods over various student and teacher segmentation networks on ADE20K.}
\label{tab: SOTA on AED20K}
\resizebox{0.8\linewidth}{!}{
\begin{tabular}{l |c c} 
\toprule[1.5pt]
Methods  & mIoU &mAcc(\%)\\
\hline
\hline
T:PSPNet        & 44.39 & 54.74\\
\hline
S:PSPnet-R18    & 29.42 & 38.48 \\
SKDS           & 31.80 & 42.25 \\
IFVD           & 32.15 & 42.53 \\
CIRKD           & 32.25 & 43.02 \\
CWD             & 33.82 & 42.41 \\
Ours           & 35.30  & 44.06 \\
\hline
S:PSPnet-R18$^\dagger$  & 17.11 & 22.99 \\
SKDS                   & 20.79 & 27.74  \\
IFVD                   & 20.75 & 27.6  \\
CIRKD                   & 22.90 & 30.68 \\
CWD                     & 25.14 & 34.13 \\
Ours                   & 26.96 & 34.13 \\
\hline
\hline
T:HRNet & 42.02 & 53.52 \\
\hline
S:HRNet18s  & 28.69 & 37.86 \\
SKDS       & 30.49 & 40.19 \\
IFVD       & 30.57 & 40.42 \\
CIRKD       & 31.34 & 41.45 \\
CWD         & 31.36 & 39.68 \\
Ours       & 32.46 & 41.92 \\
\hline
\hline
T:DeeplabV3Plus & 45.47 & 56.41 \\
\hline
S:Deeplab-MV2   & 22.38 & 31.71 \\
SKDS           & 24.65 & 35.07 \\
IFVD           & 24.53 & 35.13 \\
CIRKD       & 25.21 & 36.17 \\
CWD             & 26.89 & 35.79 \\
Ours           & 28.29 & 38.31 \\
\hline
\hline
T:ISANet    & 43.80 & 54.39 \\
\hline
S: ISANet-R18   & 27.68 & 36.92 \\
SKDS           & 28.70 & 38.51  \\
IFVD           & 29.66 & 39.80  \\
CIRKD           & 29.79 & 40.48 \\
CWD             & 34.69 & 43.05 \\
Ours           & 35.74 & 44.55 \\
\toprule[1.5pt]
\end{tabular}
}
\vspace{-1em}
\end{center}
\end{table}

\noindent\textbf{Pascal Context.}
Table \ref{tab: SOTA on Pascal Context} summarizes our results on Pascal Context validation set. Our method achieves the best performance consistently. It surpasses the best completing CWD \cite{shu2021channel} with a $0.45\%$ mIoU improvement, $0.11\%$ mIoU improvements, and $0.92\%$ mIoU improvement on PSPnet-R18, HRNet18s, and Deeplab-MV2. More segmentation results are shown in 
Figure in Supplementary.

\begin{table}[!htbp]
\begin{center}
\caption{Performance comparison with state-of-the-art distillation methods over various student and teacher segmentation networks on Pascal Context.}
\label{tab: SOTA on Pascal Context}
\resizebox{0.7\linewidth}{!}{
\setlength{\tabcolsep}{4pt}
\begin{tabular}{l |c c} 
\toprule[1.5pt]
\hline 
Methods  & mIoU &mAcc(\%)\\
\hline
\hline
T:PSPNet        & 52.47 & 63.15\\
\hline
S:PSPnet-R18    & 43.07 & 53.79 \\
SKDS           & 43.93 & 54.01 \\
IFVD           & 44.75 & 54.99 \\
CIRKD           & 44.83 & 55.3 \\
CWD             & 45.92 & 55.55 \\
Ours           & 46.37 & 56.39 \\
\hline
\hline
T:HRNet     & 51.12 & 61.39 \\
\hline
S:HRNet18s  & 40.82 & 51.70 \\
SKDS       & 42.91 & 53.63 \\
IFVD       & 43.12 & 54.03 \\
CIRKD       & 43.45 & 54.1 \\
CWD         & 45.50 & 56.01 \\
Ours       & 45.61 & 56.13 \\
\hline
\hline
T:DeeplabV3Plus & 53.20 & 64.04 \\
\hline
S:Deeplab-MV2   & 37.16 & 49.10 \\
SKDS           & 39.18 & 51.13 \\
IFVD           & 38.80 & 50.79 \\
CIRKD       & 39.99 & 52.66 \\
CWD             & 42.52 & 53.24 \\
Ours           & 43.44 & 54.29 \\
\hline
\toprule[1.5pt]
\end{tabular}
}
\vspace{-1em}
\end{center}
\end{table}

\begin{table}[!htbp]
\begin{center}
{
\caption{Performance comparison with state-of-the-art distillation methods over various student and teacher segmentation networks on Cityscapes dataset and Pascal VOC dataset.}
\label{tab: SOTA on 3 dataset}
\resizebox{0.9\linewidth}{!}{
\setlength{\tabcolsep}{7pt}
\begin{tabular}{l |c c |c c } 
\toprule[1.5pt]
\multirow{2}{*}{Methods} &
\multicolumn{2}{c|}{Cityscapes} &
\multicolumn{2}{c}{Pacal VOC}\\
\cline{2-5}
& mIoU &mAcc(\%) & mIoU &mAcc(\%)\\
\hline
\hline
T:PSPNet            & 79.74 & 86.56 		    & 78.52 & 86.11\\
\hline
S:PSPnet-R18	    & 68.99 & 75.19 		& 70.52 & 81.04 \\
SKDS               & 69.33 & 75.37 		    & 70.35 & 80.22 \\
IFVD               & 71.08 & 77.46		    & 70.92 & 81.31 \\
CIRKD              & 72.23 & 78.79          & 70.13 & 80.24 \\
CWD                & 74.29 & 80.95 	        & 73.36 & 82.63 \\
Ours               & 74.75 & 82.0 		    & 73.97 & 82.96 \\
\hline
\hline
T:HRNet     	    & 80.65 & 87.39 		    & 76.24 & 84.95 \\
\hline
S:HRNet18s  		& 73.77 & 82.89 	    & 67.47 & 78.99 \\
SKDS       		    & 74.75 & 83.23 	   & 67.58 & 79.10 \\
IFVD       		& 75.33 & 83.83 	    & 67.5 & 78.89 \\
CIRKD               & 74.63 & 83.72         & 67.36 & 79.22 \\
CWD  		      	& 75.54 & 84.08 	    & 68.39 & 79.78 \\
Ours     			& 75.93 & 84.75 	    & 69.04 & 80.37 \\
\hline
\hline
T:DeeplabV3Plus	    & 80.98 & 88.7 			    & 78.62 & 86.55 \\
\hline
S:Deeplab-MV2  	    & 70.49 & 80.11 		    & 62.56 & 80.09 \\
SKDS           	& 70.81 & 79.31 		    & 62.85 & 80.25 \\
IFVD           	& 71.82 & 80.88 	    & 67.50 & 78.89 \\
CIRKD           & 72.39 & 81.84             & 63.57 & 79.63 \\
CWD             	& 73.35 & 82.41 	    & 67.61 & 82.03 \\
Ours           	& 74.08 & 83.83 	   & 67.62 & 82.13 \\
\hline
\hline
T:ISANet        	& 80.61 & 88.29 			    & 78.46 & 87.33 \\
\hline
S: ISANet-R18   	& 71.45 & 78.65	  		    & 68.71 & 79.81 \\
SKDS           	& 70.65 & 77.53	   		    & 67.86 & 80.47  \\
IFVD           	& 70.30 & 77.79	 	    & 69.11 & 80.93  \\
CIRKD           & 72.00 & 79.32             & 69.0 & 80.83 \\
CWD             	& 71.61 & 80.02	 	    & 72.83 & 83.99 \\
Ours           	& 72.44 & 81.04	   	    & 73.17 & 84.25 \\
\toprule[1.5pt]
\end{tabular}
}
}
\end{center}
\vspace{-1em}
\end{table}

\noindent\textbf{Cityscapes.}
Table \ref{tab: SOTA on 3 dataset} lists the performance results of other state-of-the-art methods and ours on the Cityscapes dataset. We can observe that all structured knowledge distillation methods improve the student networks under the teacher's supervision. Our method achieves the best segmentation performance across various student networks. Compared with other state-of-the-art methods, we can see that our method can effectively improve the performance by $0.46\%$, $0.39\%$, $0.73\%$, $0.83\%$, respectively. Meanwhile, compared with the student, our method can improve the performance by $5.76\%$, $2.16\%$, $3.59\%$, $0.99\%$, respectively. It is also shown that our method works from natural scenes to street scenes. 
As shown in Figure in Supplementary,
our approach allows for better handling of the truck and bus categories, enabling them to be completed. In addition, the segmentation results of our method are more detailed.

\noindent\textbf{Pascal VOC.}
In Table \ref{tab: SOTA on 3 dataset},  we show the segmentation performance of various distillation methods on Pascal VOC. 
Compared with the student without distillation, we can see that our algorithm can effectively improve the performance by $3.45\%$, $1.57\%$, $5.06\%$, $4.46\%$, respectively. Meanwhile, our methods achieve the best performance on different students. We further show the qualitative segmentation results visually 
in Figure in Supplementary.

\subsection{Ablation Study}
In this subsection, we conduct experiments to explore the effectiveness of our methods under different knowledge distillation settings. All ablation experiments are carried out on the ADE20k dataset. We choose PSPnet-R101 as the teacher and PSPnet-R18 as the student.

\noindent\textbf{Ablation study of different equations.}
As shown in Table \ref{tab: Ablation Study}, compared with $\ell_{PD}$, $\ell_{aug}^{PD}$ introduces infinite samples with $1.09\%$ improvement.
Meanwhile, compared with $\ell_{CWD}$, $\ell_{aug}^{CWD}$ also introduces infinite samples with $1.48\%$ improvement. The introduction of infinite samples allows for a more accurate simulation of the decision boundary of the teacher model.

\begin{table}[ht]
\centering
\setlength{\tabcolsep}{15pt}
\caption{
Ablation study of different equations on ADE20K.
Introducing infinite samples in $\ell_{PD}$ and $\ell_{CWD}$ both bring improvements.
}
\label{tab: Ablation Study}
\begin{tabular}{l |c c} 
\hline 
\toprule[0.5pt]
Methods  & mIoU &mAcc(\%)\\
\hline
\hline
$\ell_{PD}$ &31.75 &42.23 \\
$\ell_{aug}^{PD}$ &32.84 &42.33 \\
\hline
$\ell_{CWD}$ &33.82 &42.41 \\
$\ell_{aug}^{CWD}$ &35.30 &44.06 \\
\hline
\toprule[0.5pt]
\end{tabular}
\vspace{-1em}
\end{table}

\begin{table}[!htbp]
\begin{center}
\caption{Experiment for different $\lambda$. $\lambda$ is introduced in Equation \ref{eq: gaus}. And $1.0$ is the best choice for $\lambda$.}
\label{tab: Experiment for different ratio.}
\begin{tabular}{l |c c} 
\hline 
\toprule[0.5pt]
Ratio  & mIoU &mAcc(\%)\\
\hline
0.5 &35.06 &43.91 \\
1.0 &35.30 &44.06 \\
1.5 &34.63 &42.97 \\
2.5 &31.95 &39.04 \\
\hline
\toprule[0.5pt]
\end{tabular}
\end{center}
\vspace{-1em}
\end{table}

\begin{figure}[!htbp]
\centering
\includegraphics[width=0.8\linewidth]{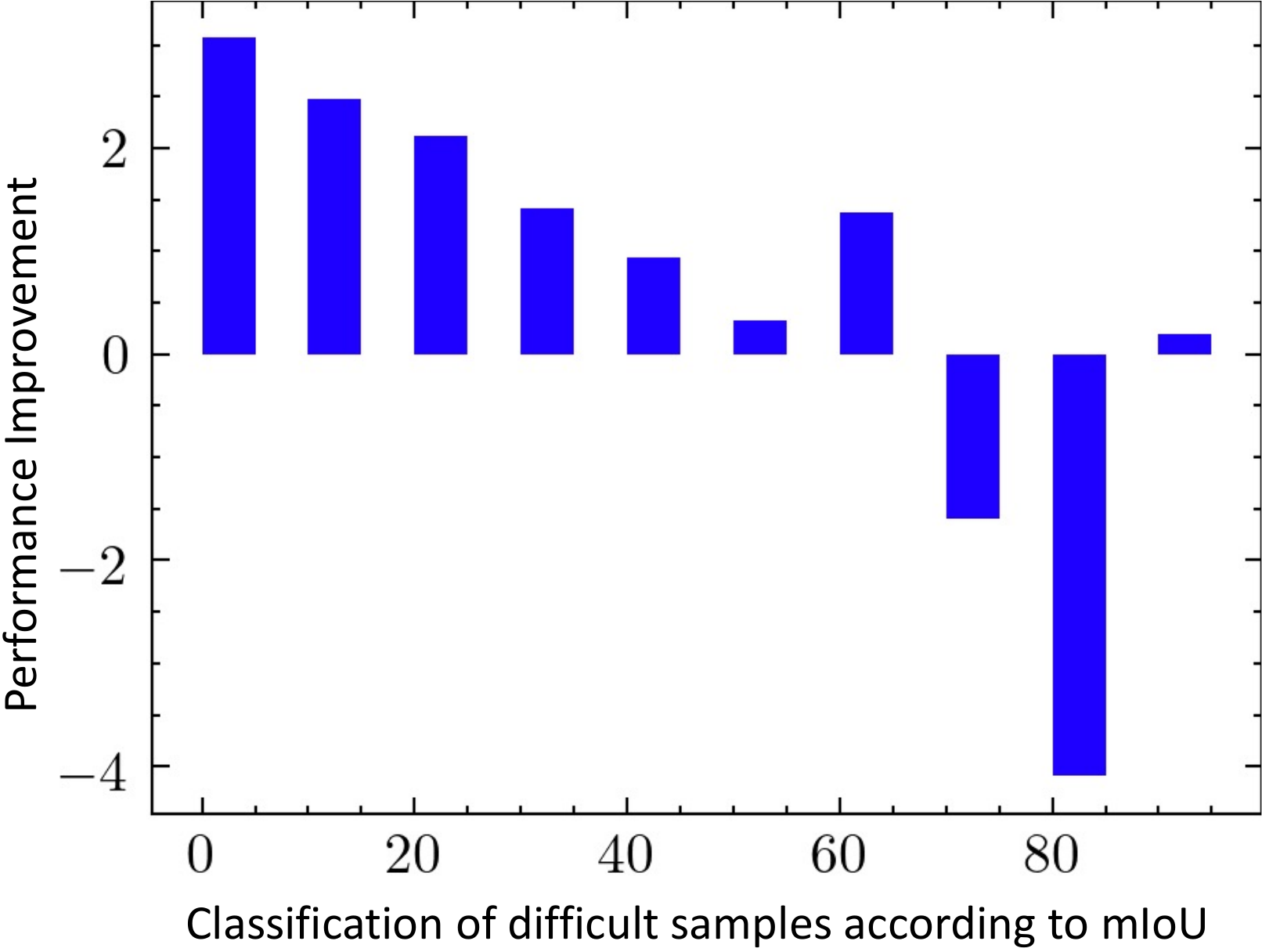}
\caption{Analysis of samples with different difficulties. All categories are divided into difficult and easy samples according to the performance. The abscissa represents the difficulty of the category, which is becoming simpler and simpler from left to right. The ordinate represents a performance improvement.}
\label{fig: hard example}
\vspace{-1em}
\end{figure}

\noindent\textbf{Ablation study of $\lambda$.} As shown in $\ell_{aug}^{CWD}$, there has $\lambda$ to adjust the ratio. However, since covariance is tunneled training process by data set statistics, it is less stable in the early stage of training. For stabilize training, $\lambda$ uses the cosine update method. As shown in Table \ref{tab: Experiment for different ratio.} , we investigate the impact of $\lambda$ in our FAKD, and $\lambda = 1.0$ is the best choice.

\begin{figure}[!htbp]
  \centering
  \includegraphics[width=0.86\linewidth]{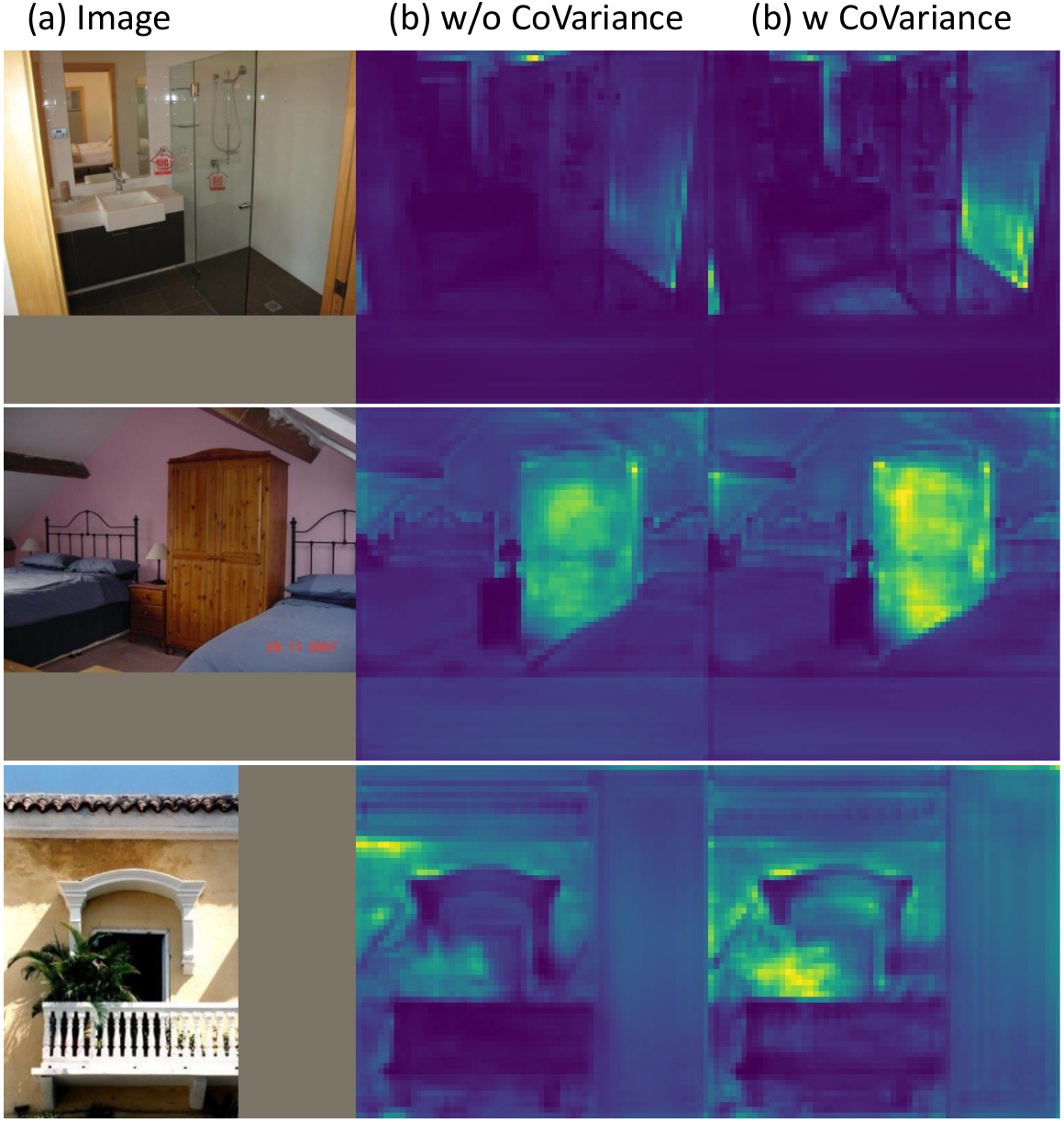}
  \caption{Visual of samples with or without co-variance. By augmenting the feature with co-variance, the activation of the meaningful region is strengthen.  }
  \label{fig: feature visual}
\vspace{-1em}
\end{figure}

\begin{figure}[ht]
\begin{center}
    \subfigure[ADE20K]{\includegraphics[scale=0.49]{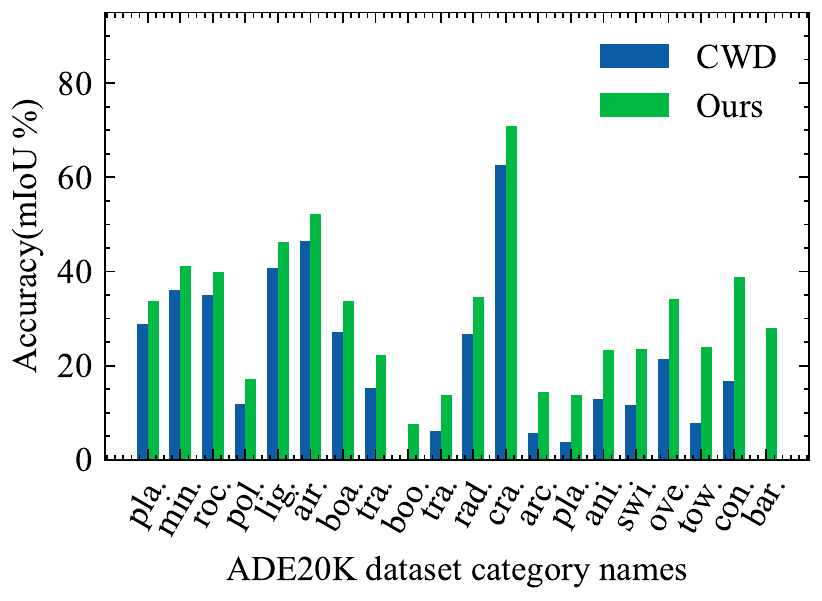}}
    \subfigure[Cityscapes]{\includegraphics[scale=0.49]{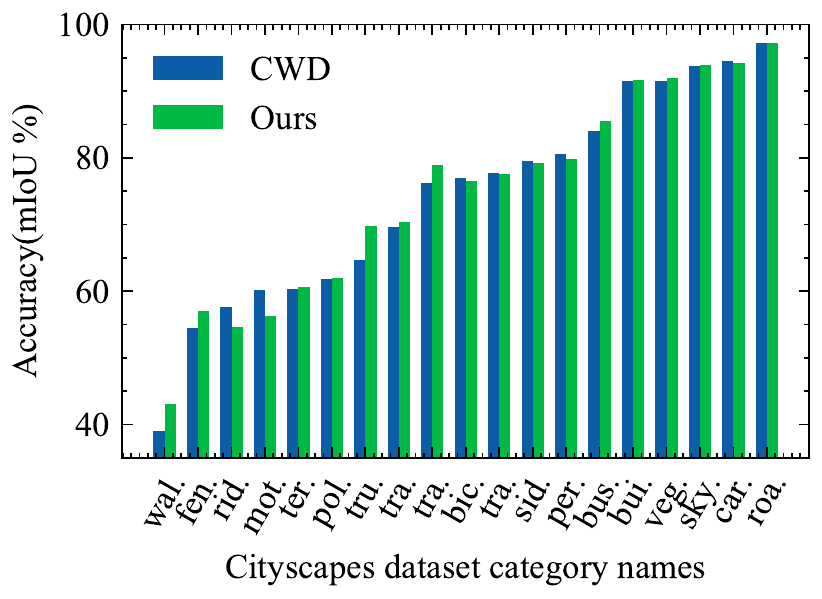}} 
    \quad
    
	\subfigure[Pascal Cotext]{\includegraphics[scale=0.49]{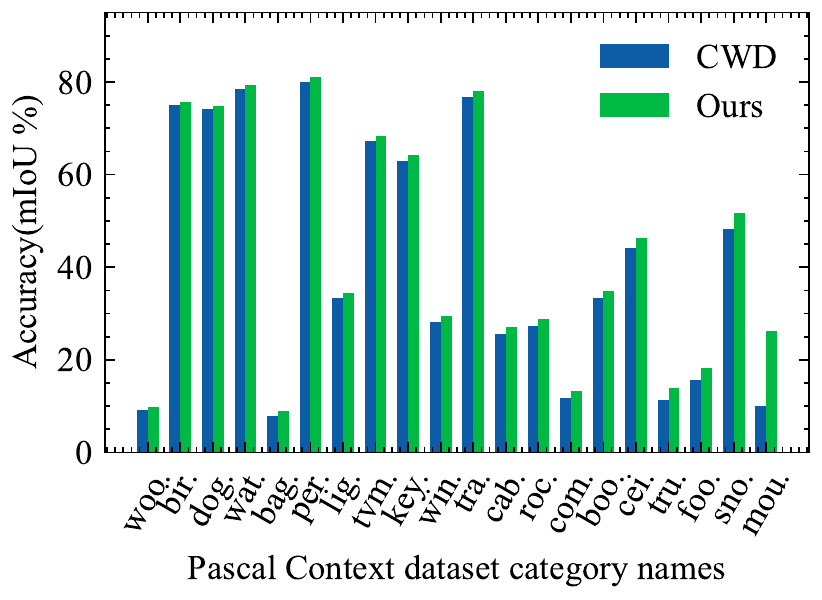}}
	\subfigure[Pascal VOC]{\includegraphics[scale=0.49]{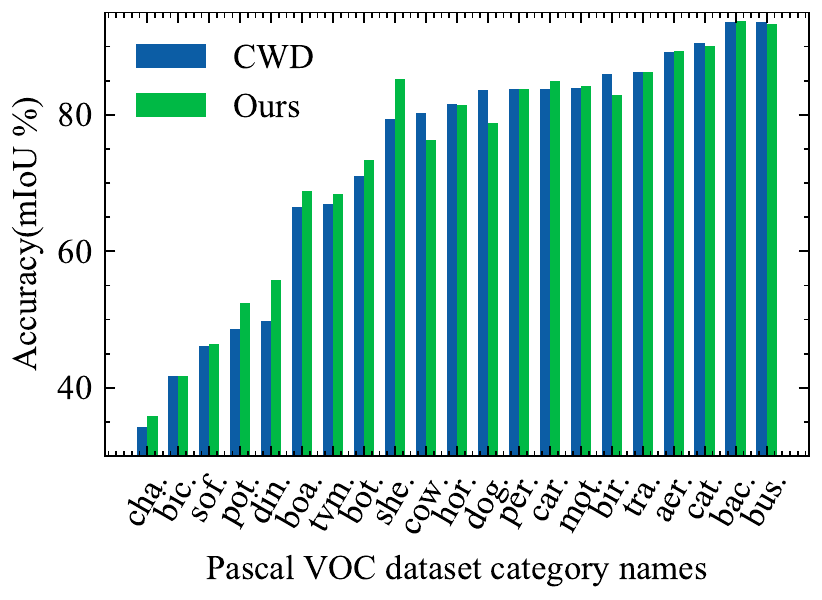}} 
\end{center}
\caption{Illustrations of the effectiveness of our methods in terms of class IoU scores using the network PSPnet-ResNet18. Both CWD and our method are helpful for improving the performance especially for the hard classes with low IoU scores. The improvement from our method is more significant for hard objects.}
\label{fig:each class}
\vspace{-1em}
\end{figure}

\noindent\textbf{Ablation study of hard example mining.} As shown in Figure \ref{fig: hard example}, the item with $\Sigma$ in $\ell_{aug}^{CWD}$ and $\ell_{aug}^{PD}$ is reflected as a dynamically adjusted complex sample. So we divided the different difficult samples and analyzed the performance improvement for samples of different difficulties. From the results, it can be seen that our method can effectively improve the problematic samples. Also, we visualized the feature maps as Illustrated in Figure \ref{fig: feature visual}. It can also be seen from the feature map that adding our method enables the feature map to highlight some problematic regions more. Each class IoU score is shown in Figure \ref{fig:each class}.

\section{Conclusion}
\label{sec:Conclusion}
This paper presents a novel feature augmented knowledge distillation (FAKD) method for semantic segmentation to increase the training samples for the student network in the feature space. Our method augments feature maps and constrains all augmented features with the teacher network. Experiments on four public segmentation datasets demonstrate the effectiveness of our FAKD. Our method also demonstrates superiority on long-tail problems. The performance for classes with less training samples has been improved with a larger margin. We hope our work inspires future research exploring semantic feature augmentation for knowledge distillation.


{\small
\bibliography{aaai23}
}


\end{document}